\documentclass[runningheads]{llncs}
\usepackage{graphicx}

\usepackage{tikz}
\usepackage{comment}
\usepackage{amsmath,amssymb} 
\usepackage{color}

\usepackage{multirow}
\usepackage{axessibility}  

\usepackage{epsfig}
\usepackage[labelformat=simple]{subfig}

\usepackage{algorithm}      
\usepackage[noend]{algpseudocode} 
\usepackage{url}

\newcommand{\algrule}[1][.2pt]{\par\vskip.1\baselineskip\hrule height #1\par\vskip.1\baselineskip}
\algdef{SE}[DOWHILE]{Do}{doWhile}{\algorithmicdo}[1]{\algorithmicwhile\ #1}%

\begin{document}
\pagestyle{headings}
\mainmatter
\def\ECCVSubNumber{83}  

\title{Self-paced learning to improve text row detection in historical documents\\with missing labels} 


\titlerunning{Self-paced learning to improve text row detection in historical documents}
%
\author{Mihaela G\u{a}man \and
Lida Ghadamiyan \and\\
{Radu Tudor} Ionescu\index{Ionescu, Radu Tudor}\inst{*} \and Marius Popescu}
\authorrunning{M. G\u{a}man et al.}
%
\institute{Department of Computer Science\\
University of Bucharest\\
14 Academiei, Bucharest, Romania\\
*\email{raducu.ionescu@gmail.com}}

\maketitle

\begin{abstract}
An important preliminary step of optical character recognition systems is the detection of text rows. To address this task in the context of historical data with missing labels, we propose a self-paced learning algorithm capable of improving the row detection performance. We conjecture that pages with more ground-truth bounding boxes are less likely to have missing annotations. Based on this hypothesis, we sort the training examples in descending order with respect to the number of ground-truth bounding boxes, and organize them into $k$ batches. Using our self-paced learning method, we train a row detector over $k$ iterations, progressively adding batches with less ground-truth annotations. At each iteration, we combine the ground-truth bounding boxes with pseudo-bounding boxes (bounding boxes predicted by the model itself) using non-maximum suppression, and we include the resulting annotations at the next training iteration. We demonstrate that our self-paced learning strategy brings significant performance gains on two data sets of historical documents, improving the average precision of YOLOv4 with more than $12\%$ on one data set and $39\%$ on the other.
\end{abstract}

\section{Introduction}
\label{sec:intro}

Automatically processing historical documents to extract and index useful information in digital databases is an important step towards preserving our cultural heritage \cite{Lombardi-JI-2020}. This process is commonly based on storing the information as images. However, performing optical character recognition (OCR) on the scanned documents brings major benefits regarding the identification, storage and retrieval of information \cite{Martinek-NCA-2020,Martinek-ICDAR-2019,Neudecker-DATeCH-2019,Nunamaker-ICIP-2016}.

\begin{figure*}[!t]
\begin{center}
\centering
\includegraphics[width=1.0\linewidth]{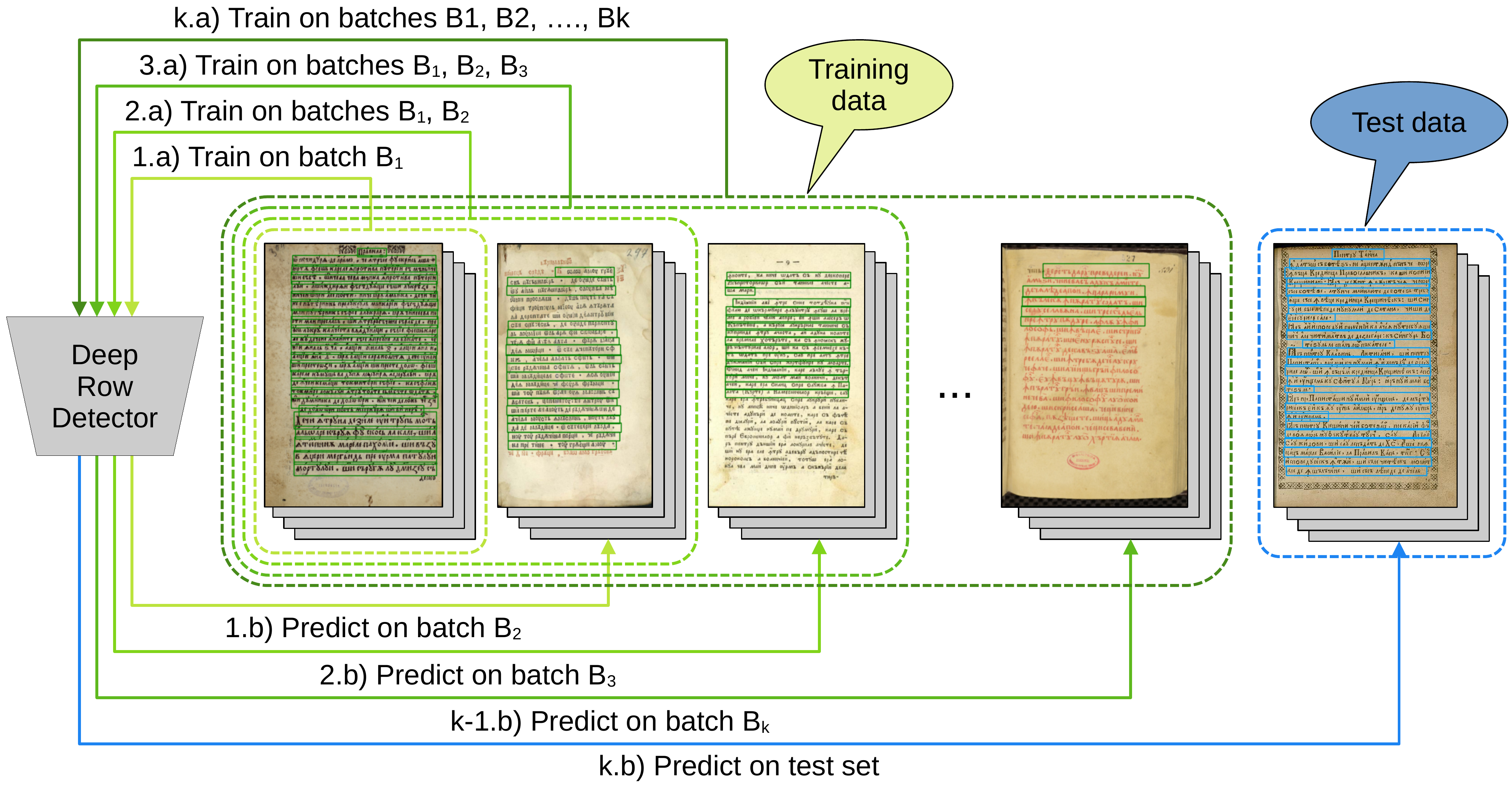}
\vspace{-0.2cm}
\caption{Our self-paced learning algorithm for row detection in historical documents with missing bounding box annotations. Training examples are sorted into $k$ batches having less and less ground-truth annotations. Best viewed in color.}
\label{fig_pipeline}
\end{center}
\end{figure*}

Prior to the recognition of handwritten or printed characters, an important step of OCR systems is the detection of text lines (rows) \cite{Gruning-IJDAR-2019,Mechi-ICDAR-2019,Mechi-ICPR-2021,Melnikov-DAS-2020}. Our work is particularly focused on the text row detection task under a difficult setting, where the training data has missing labels (bounding boxes). To improve detection performance on this challenging task, we propose a self-paced learning algorithm that gradually adds pseudo-labels during training. Our algorithm is based on the following hypothesis: pages with more ground-truth bounding boxes are less likely to have missing annotations. We are certain that this hypothesis holds if the number of text rows per page is constant. However, in practice, pages may come from different books having various formats. Moreover, some pages at the end of book chapters may contain only a few rows. Even if such examples are present, we conjecture that our hypothesis holds in a sufficiently large number of cases to be used as a function to schedule the learning process. We thus propose to organize the training examples into $k$ batches, such that the first batch contains the most reliable pages (with the highest number of bounding boxes per page). The subsequent batches have gradually less annotations. The training starts on the first batch and the resulting model is applied on the second batch to enrich it with pseudo-labels (bounding boxes predicted by the model itself). One by one, batches with ground-truth and pseudo-labels are gradually added during training, until the model gets to see all training data. Our self-paced learning process is also illustrated in Figure~\ref{fig_pipeline}.

We conduct text row detection experiments on two data sets of historical documents, comparing the state-of-the-art YOLOv4 \cite{bochkovskiy2020yolov4} detector with a version of YOLOv4 trained under our self-paced learning regime. The empirical results show that our self-paced learning algorithm introduces significant performance gains on both benchmarks. 

\paragraph{\bf Contribution.}
In summary, our contribution is twofold:
\begin{itemize}
\item We propose a novel self-paced learning algorithm for line detection which considers the training examples in the decreasing order of the number of ground-truth bounding boxes, alleviating the problem of missing labels by introducing pseudo-labels given by the detector.
\item We present empirical results demonstrating that our self-paced learning regime brings significant improvements in text row detection on two data sets, under a fair yet challenging evaluation setting that ensures source (document) separation (and, on one data set, even domain shift from the Cyrillic script to the Latin script) across training and test pages.
\end{itemize}

\section{Related Work}
\label{sec:relatedart}

\paragraph{\bf Self-paced learning.}
Humans hold the ability to guide themselves through the learning process, being capable of learning new concepts at their own pace, without requiring instructions from a teacher. Indeed, learners often choose what, when and how long to study, implicitly using a self-paced curriculum learning process. This learning process inspired researchers to propose and study artificial neural networks based on self-paced learning (SPL) \cite{jiang2015self,kumar2010self}. Researchers have proposed a broad range of SPL strategies for various computer vision \cite{jiang2015self,lin2020pixel,Soviany-CVIU-2021} and pattern recognition tasks \cite{ristea2021self,zheng2020unsupervised,zhou2020self}, demonstrating performance gains over the standard supervised learning method. Although SPL can be viewed as a particularization of curriculum learning \cite{Soviany-A-2021}, where the complexity of the samples is estimated by the learner, we consider the broader domain of curriculum learning as less relevant to our work and choose not to cover it in this section. 

To the best of our knowledge, we are the first to employ self-paced learning for text row detection.

\paragraph{\bf Object detection.} One way to approach text row detection is to employ a state-of-the-art deep object detector. We distinguish two types of deep object detectors in the recent literature. On the one hand, there are two-stage object detectors, e.g.~Faster R-CNN \cite{Ren-NIPS-2015}, that employ a region proposal network to generate object proposals in the first stage. The proposals are subsequently refined in the second detection stage by the final model. On the other hand, there are one-stage detectors, e.g.~YOLO \cite{Redmon-CVPR-2016} and SSD \cite{Liu-ECCV-2016},  capable of detecting objects in a single run. YOLO is one of the most popular object detectors, being continuously improved by the community with new versions such as YOLOv3 \cite{redmon2018yolov3} and
YOLOv4 \cite{bochkovskiy2020yolov4}. In our self-paced learning algorithm, we employ the recent YOLOv4 as the underlying detector.

\paragraph{\bf Row detection.}
Text line extraction is an important initial step in the analysis of document images \cite{Gruning-IJDAR-2019}. Historical writings pose a variety of challenges in the detection of text lines, from multiple different editorial styles to noise such as yellow pages or ink stains, and even to significant physical degradation \cite{Mechi-ICDAR-2019,Mechi-ICPR-2021}. Due to such particularities, a reduction in performance is encountered in the processing of ancient text. Thus, even recent works consider the problem of text line detection open and yet to be solved \cite{Gruning-IJDAR-2019,Mechi-ICDAR-2019,Mechi-ICPR-2021}.
In the past few years, researchers continued to make progress in the matter of text line detection in historical documents. 
We hereby highlight a few recent efforts which successfully employed deep learning to solve this task. Some approaches \cite{Mechi-ICDAR-2019,Mechi-ICPR-2021} perform text line segmentation, relying on the U-Net architecture or its variations, such as ARU-Net \cite{Gruning-IJDAR-2019}. Hybrid architectures, combining U-Net with traditional image processing techniques worked best in some recent works \cite{Gruning-IJDAR-2019,Mechi-ICPR-2021}. A trade-off between accuracy and computational efficiency is studied in \cite{Melnikov-DAS-2020}. Here, the authors introduce a lightweight CNN suitable for mobile devices.

To the best of our knowledge, we are the first to study text line detection under missing labels.

\section{Method}
\label{sec:method}

Assuming that all pages have about the same number of text rows, we conjecture that pages with more bounding boxes are less likely to have missing annotations. Our self-paced learning algorithm relies on this conjecture to organize and present the training data in a meaningful order, which allows us to learn a more robust detector. The steps of our self-paced learning method are formalized in Algorithm \ref{alg_1}.

\newlength{\textfloatsepsave} \setlength{\textfloatsepsave}{\textfloatsep}

\begin{algorithm}[!t]
\caption{Self-paced learning for row detection}\label{alg_1}
\textbf{Input: }{$X=\{(x_i,T_{x_i})\;|\;x_i \in \mathbb{R}^{h\times w \times c}, T_{x_i} = \{t_{x_i}^j\;|\;t_{x_i}^j \in \mathbb{R}^5\}, \forall i \in \{1,2,...,n\} \}$ - a training set of $n$ images and the associated bounding boxes (each bounding box $t_{x_i}^j$ is represented by its four coordinates and a confidence score); $Y=\{y_i\;|\;y_i \in \mathbb{R}^{h\times w \times c}, \forall i \in \{1,2,...,m\} \}$ - a test set of $m$ images; $k$ - the number of training batches; $\eta$ - a learning rate; $\mathcal{L}$ - a loss function; $p$ - IoU threshold for the NMS procedure.}\\
\textbf{Notations: }{$f$ - a deep row detector; $\theta$ - the weights of the detector $f$; \emph{nms} - a function that applies non-maximum suppression; $\mathcal{N}(0, \Sigma)$ - the normal distribution of mean $0$ and standard deviation $\Sigma$; $\mathcal{U}(S)$ - the uniform distribution over the set $S$.}\\
\textbf{Initialization: }{$\theta^{(0)} \sim \mathcal{N}(0, \Sigma)$}\\
\textbf{Output: }{$\mathcal{P} = \{ P_{y_1},P_{y_2}, ..., P_{y_m} \}$ - the bounding boxes predicted for $y_i \in Y, \forall i \in \{1,2,...,m\}$.}
\vspace{0.2em}
\algrule
\vspace{0.2em}
\textbf{Stage 1:}{ Training data organization}
\vspace{0.2em}
\algrule
\vspace{0.2em}
\begin{algorithmic}[1]
\State sort and split $X$ into $B_1, B_2, ..., B_k$ such that:\\
\hspace{1em} $X = \bigcup_{i=1}^k B_i$\\
\hspace{1em} $B_i \cap B_j = \emptyset, \forall i,j \in \{1,2,...,k\}, i \neq j$\\
\hspace{1em} $|B_i|-|B_j| \leq 1, \forall i,j \in \{1,2,...,k\}$\\
\hspace{1em} $\forall x \in B_i, z \in B_j, i < j, |T_x| \geq |T_z| $
\vspace{0.2em}
\algrule
\vspace{0.2em}
\noindent\hspace{-2em}
\textbf{Stage 2:}{ Self-paced learning}
\vspace{0.2em}
\algrule
\vspace{0.2em}
\State $X' \gets \emptyset$
\For{$i \gets 1$ to $k$}
    \State $X' \gets X' \cup B_i$
    \State $t \gets 0$
    \While{converge criterion not met}
     \State $X^{(t)} \gets$ mini-batch $\sim \mathcal{U}(X')$
     \State $\theta^{(t+1)} \gets \theta^{(t)} - \eta^{(t)} \nabla{\mathcal{L}\left(\theta^{(t)}, X^{(t)}\right)}$
     \State $t \gets t + 1$
    \EndWhile
    
    \If{$i+1 \leq k$}
        \For{$x \in B_{i+1}$}
            \State $P_x \gets f(\theta,x)$
            \State $T_x \gets T_x \cup P_x$
            \State $T_x \gets nms(T_x, p)$
        \EndFor
    \EndIf
\EndFor
\vspace{0.2em}
\algrule
\vspace{0.2em}
\noindent\hspace{-2em}
\textbf{Stage 3:}{ Prediction on test set}
\vspace{0.2em}
\algrule
\vspace{0.2em}
\For{$i \gets 1$ to $m$}
    \State $P_{y_i} \gets f(\theta, y_i)$
\EndFor
\end{algorithmic}
\end{algorithm}
\setlength{\textfloatsep}{7pt}

In the first stage (steps 1-5), we split the training set $X$ into a set of $k$ disjoint (according to step 3) and equally-sized (according to step 4) batches, denoted as $B_1, B_2$, ..., $B_k$. Before splitting the data, we sort the training images with respect to the number of bounding boxes, in descending order. Hence, for any two images $x \in B_i$ and $z \in B_j$, where $i< j$, the number of bounding boxes associated to $x$ is higher or equal to the number of bounding boxes associated to $z$ (according to step 5).

Once the training data is organized into batches, we proceed with the second stage (steps 6-18). The self-paced learning procedure is carried out for $k$ iterations (step 7), alternating between two loops at each iteration $i$. In the first loop (steps 10-13), the model $f$ is trained with stochastic gradient descent on batches $B_1, B_2, ..., B_i$, using as target ground-truth bounding boxes as well as pseudo-bounding boxes, which are predicted at the previous iterations $1, 2, ..., i-1$. In the second loop (steps 15-18), we iterate through the training images in batch $B_{i+1}$ and apply the detector $f$ (step 16) on each sample $x$ to predict the bounding boxes $P_x$. The predicted bounding boxes are added to the set of ground-truth boxes $T_x$ (step 17). Then, we apply non-maximum suppression (step 18) to eliminate pseudo-bounding boxes that overlap with the ground-truth ones. We hereby note that we assign a confidence score of $1$ to all ground-truth bounding boxes, ensuring that these never get eliminated. In other words, the NMS procedure at step 18 can only eliminate pseudo-annotations.

In the third stage (steps 19-20), we employ the final model to detect text rows in the test images. The resulting bounding boxes represent the final output of our model.

Aside from the common hyperparameters that configure the standard optimization process based on stochastic gradient descent, our algorithm requires two additional hyperparameters, $k$ and $p$. The former hyperparameter ($k$) determines the number of batches, while the second hyperparameter ($p$) specifies the threshold which decides when to suppress pseudo-bounding boxes.

\setlength{\textfloatsep}{\textfloatsepsave}

\section{Experiments}
\label{sec_experiments}

\subsection{Data Sets}

\paragraph{\bf ROCC.} The Romanian Old Cyrillic Corpus (ROCC) represents a collection of scanned historical documents, spanning a time frame of more than three centuries \cite{Cristea-INCOMA-2020,Cristea-ConsILR-2022}. 
The corpus provided by Cristea et al.~\cite{Cristea-INCOMA-2020} consists of 367 scanned document pages, with a total of 6418 annotated text lines. 
We select the training and test pages from different distributions, i.e.~distinct books, ensuring a fair and unbiased evaluation. We split the data into a training set consisting of 332 pages (with 5715 annotated bounding boxes) and a test set of 35 pages (with 492 bounding boxes). Both training and test splits have a significant amount of missing bounding boxes (rows that are not annotated). For a correct evaluation, we manually annotate the missing lines in the test set, 
reaching a total of 703 bounding boxes for the final test set.

\paragraph{\bf cBAD.} The cBAD competition \cite{Diem-ICDAR-2017} for baseline detection featured a corpus formed of 2035 pages of historical documents written in the Latin script. To test the robustness of our self-paced learning method, we randomly select 52 images from the cBAD data set to serve as an out-of-domain test set (the training is conducted on ROCC). We manually annotate the selected pages with $1542$ bounding boxes representing text rows.

\subsection{Evaluation Setup}

\paragraph{\bf Evaluation metrics.} As evaluation measures, we report the average precision (AP) at an Intersection over Union (IoU) threshold of $0.5$, as well as the mean IoU.

\paragraph{\bf Baselines.} As the first baseline, we consider the YOLOv4 \cite{bochkovskiy2020yolov4} based on the conventional training procedure. We also add a baseline based on self-paced learning with 5 batches, which takes the examples in a random order, in contrast to our approach that sorts the examples based on the number of bounding boxes.

\begin{table}[!t]
\setlength\tabcolsep{4.2pt}
\caption{Baseline YOLOv4 versus our YOLOv4 based on self-paced learning (SPL). AP and mean IoU scores (in \%) are reported on two data sets: ROCC and cBAD. For ROCC, a baseline SPL regime that takes examples in a random order is included. Best scores on each data set are highlighted in bold.}
\label{tab_Results}
\begin{center}
\begin{tabular}{|c|l|c|c|c|}
\hline
{\bf Data Set} & {\bf Model} & {\bf Iteration}    & {\bf AP}    & {\bf Mean IoU}\\
\hline
\hline
\multirow{11}{*}{\rotatebox{90}{ROCC}}&YOLOv4 (baseline)     & -	& 81.55	                & 70.60   \\ 
\cline{2-5}
&\multirow{5}{*}{YOLOv4 + SPL (random)} &1     & 70.06	            & 57.53   \\ 
& &2    & 74.86	            & 55.53   \\ 
& &3    & 75.95	            & 54.20   \\ 
& &4    & 75.16	            & 58.69   \\ 
\cline{3-5}
& & 5 (final)    & 77.56	& 57.77    \\
\cline{2-5}
&\multirow{5}{*}{YOLOv4 + SPL (ours)} &1     & 72.43	            & 65.80   \\ 
& &2    & 87.22	            & 66.97   \\ 
& &3    & 88.05	            & 67.06   \\ 
& &4    & 89.86	            & 69.14   \\ 
\cline{3-5}
& & 5 (final)    & \textbf{93.73}	& \textbf{75.25}    \\
\hline
\hline
\multirow{6}{*}{\rotatebox{90}{cBAD}}&YOLOv4 (baseline)     & -		        & 35.37	            & 61.30  \\
\cline{2-5}
&\multirow{5}{*}{YOLOv4 + SPL (ours)} &1		        & 22.52	            & 0.00   \\
& &2		        & 54.81	            & 56.62   \\
& &3		        & 63.14	            & 63.00   \\
& &4		        & 63.72	            & 58.96   \\
\cline{3-5}
& & 5 (final)		        & \textbf{74.57}	& \textbf{67.52}    \\
\hline
\end{tabular}
\end{center}
\vskip -0.14in
\end{table}

\paragraph{\bf Hyperparameter choices.} We employ the official YOLOv4 implementation from \cite{bochkovskiy2020yolov4}, selecting CSPDarknet53 as backbone. We use mini-batches of $4$ samples. We employ the Adam optimizer with an initial learning rate of $10^{-3}$, leaving the other hyperparameters to their default values. All models are trained for a maximum of $2000$ epochs using early stopping. For the self-paced learning strategy, we use $k=5$ batches and perform $400$ epochs with each new batch. The IoU threshold for the NMS procedure is set to $p=0.5$.

\begin{figure}[!th]
\begin{center}
\centering
\includegraphics[width=0.75\linewidth]{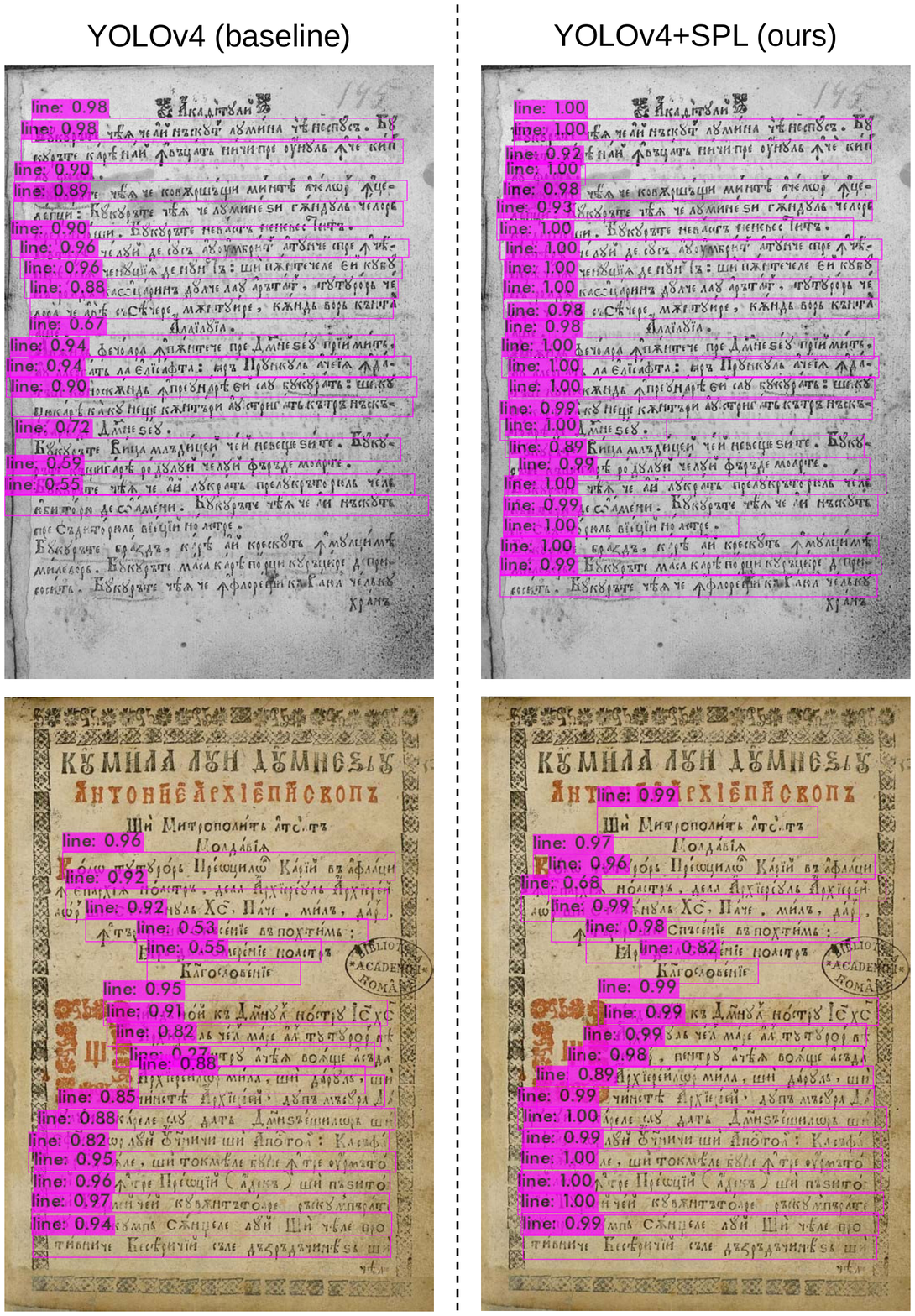}
\vspace{-0.2cm}
\caption{Detections of YOLOv4 based on the standard (baseline) training regime versus our self-paced learning (SPL) regime. Best viewed in color.}
\label{fig_examples}
\end{center}
\vspace{-0.6cm}
\end{figure}

\subsection{Results}

We present the results on ROCC and cBAD in Table~\ref{tab_Results}. First, we observe that, when the model is trained with either SPL strategy on the first batch only ($20\%$ of the training data), the amount of training data seems insufficient for the model to generalize well and compete with the baseline YOLOv4. Both SPL alternatives reach better performance levels once the whole training data is used in the learning process. However, the two SPL strategies improve with each batch at completely different paces. The baseline SPL approach based on randomly ordered examples does not even surpass the YOLOv4 based on conventional training. In contrast, once the model gets to learn at least two training batches with our SPL strategy, our training regime already surpasses the baseline, showing that it is indeed important to sort the examples according to the number of bounding boxes. Moreover, this result indicates that around $60\%$ of the training data is rather harmful for the baseline, due to the missing labels. In general, with each iteration of our SPL strategy, the inclusion of a new batch of images having both ground-truth and pseudo-labels brings consistent performance gains. Our best results are obtained after the last iteration, once the model gets to see the entire training data. We consider that our final improvements ($+12.18\%$ on ROCC and $+39.20\%$ on cBAD) are remarkable. We conclude that our conjecture is validated by the experiments, and that our self-paced learning method is extremely useful in improving text row detection with missing labels. We also underline that a basic SPL regime is not equally effective.

In Figure~\ref{fig_examples}, we illustrate typical examples of row detections given by the baseline YOLOv4 and the YOLOv4 based on our self-paced learning strategy. We observe that our model detects more rows, while the baseline model sometimes tends to detect two rows in a single bounding box.

\section{Conclusion}
\label{sec:conclusion}

In this paper, we proposed a self-paced learning algorithm for text row detection in historical documents with missing annotations. Our algorithm is based on the hypothesis that pages with more bounding boxes are more reliable, and thus, can be used sooner during training. Our hypothesis is supported by the empirical results reported on two data sets, which demonstrate significant improvements (between $12\%$ and $40\%$) in terms of AP.
In future work, we aim to extend the applicability of our self-paced learning method to other detection tasks which suffer from the problem of missing labels.

\paragraph{\bf Acknowledgment.}
This work has been carried out with the financial support of UEFISCDI, within the project PN-III-P2-2.1-PED-2019-3952 entitled ``Artificial Intelligence Models (Deep Learning) Applied in the Analysis of Old Romanian Language (DeLORo – Deep Learning for Old Romanian)''. This work was also supported by a grant of the Romanian Ministry of Education and Research, CNCS - UEFISCDI, project number PN-III-P1-1.1-TE-2019-0235, within PNCDI III. Authors are alphabetically ordered.

\bibliographystyle{splncs04}
\bibliography{references}
\end{document}